# The Cost of Troubleshooting Cost Clusters with Inside Information

**Thorsten J. Ottosen**   **Finn V. Jensen**
Department of Computer Science
Aalborg University
Selma Lagerlöfs Vej 300, DK-9220 Aalborg Øst, Denmark
E-mail: nesotto@cs.aau.dk, fvj@cs.aau.dk

## Abstract

Decision theoretical troubleshooting is about minimizing the expected cost of solving a certain problem like repairing a complicated man-made device. In this paper we consider situations where you have to take apart some of the device to get access to certain clusters and actions. Specifically, we investigate troubleshooting with independent actions in a tree of clusters where actions inside a cluster cannot be performed before the cluster is opened. The problem is non-trivial because there is a cost associated with opening and closing a cluster. Troubleshooting with independent actions and no clusters can be solved in $O(n \cdot \lg n)$ time ($n$ being the number of actions) by the well-known "P-over-C" algorithm due to Kadane and Simon, but an efficient and optimal algorithm for a tree cluster model has not yet been found. In this paper we describe a "bottom-up P-over-C" $O(n \cdot \lg n)$ time algorithm and show that it is optimal when the clusters do not need to be closed to test whether the actions solved the problem.

## 1 INTRODUCTION

In decision theoretical troubleshooting we are faced with a problem that needs to be solved by applying solution actions and by posing questions that gather information about the problem. The premise is that after each action we can cost free observe whether the problem was solved. The domain is assumed to be uncertain, that is, solution actions may be imperfect and information might be non-conclusive. Given a model that describes the uncertainty and the cost of actions and questions, the goal is to compute a strategy for solving the problem with the lowest expected cost.

If the model has the following assumptions:

(a) the problem is due to a single fault,
(b) different actions address different faults,
(c) costs do not depend on the previous history, and
(d) there are no questions,

then the problem is solvable in $O(n \cdot \lg n)$ time where $n$ is the number of actions. This algorithm is the well-known "P-over-C" algorithm by (Kadane and Simon, 1977) which was first brought into a troubleshooting context by (Kalagnanam and Henrion, 1990). Furthermore, if any of the above assumptions are relaxed without restrictions, the problem becomes NP-hard (Vomlelová, 2003). If assumption (a) is replaced with an assumption about multiple independent faults, an $O(n \cdot \lg n)$ P-over-C-like algorithm also exists (Srinivas, 1995). Troubleshooting without assumption (b) can also be somewhat simplified due to the dependency set algorithm of (Koca and Bilgic, 2004).

(Langseth and Jensen, 2001) proposed to relax assumption (c) slightly by considering a model where the actions can be partitioned into a flat set of so-called *cost clusters* (see Figure 1). The idea is that in order to access an action in a bottom level cluster $\mathcal{K}_i$, you need to pay an additional cost $C_i^o$ and to close the cluster you have to pay an additional cost $C_i^c$. Thereby it is possible to model e.g. the repair of complex man-made devices where you need to take apart some of the equipment to perform certain actions. If we can determine whether an action has solved the problem without assembling the cluster first, Langseth and Jensen said that the cluster has *inside information*; otherwise the cluster is *without inside information*. They furthermore describe heuristics for both problems. In this paper we present a proof of correctness of their algorithm for the problem with inside information. We furthermore extend the model to a tree of clusters and give an $O(n \cdot \lg n)$ time algorithm that is proved optimal. (Warnquist et al., 2008) describe a slightly more general cost cluster framework, but they do not address the issue of finding an efficient algorithm.

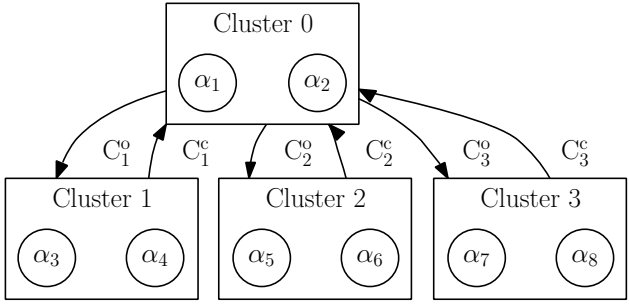

Figure 1: Example of the flat cost cluster model. To open a cluster $\mathcal{K}_i$ we pay the cost $C_i^o$, and to close a cluster we pay the cost $C_i^c$.

## 2 PRELIMINARIES

In this paper we shall examine troubleshooting problems where the following model parameters are given. $\mathcal{F} = \{f_1, \ldots, f_m\}$ is a *set of faults* describing the possible causes to the problem. For each fault $f \in \mathcal{F}$, we have a probability $P(f)$ describing how likely it is that f is present when troubleshooting begins. $\mathcal{A} = \{\alpha_1, \ldots, \alpha_n\}$ is a *set of actions* that can potentially solve the problem. Each action $\alpha$ has two possible outcomes, namely "$\alpha = yes$" (the problem was fixed) and "$\alpha = no$" (the action failed to fix the problem). Each action $\alpha$ has a positive *cost* $C_\alpha$ describing the resources required to perform the action. Finally, each action has an associated *success probability* $P(\alpha = yes|f)$, the probability of solving the problem by performing the action when f is present.

The set of actions $\mathcal{A}$ can be partitioned into $\ell+1$ clusters $\mathcal{K}, \mathcal{K}_1, \ldots, \mathcal{K}_\ell$ where cluster $\mathcal{K}$ is the *top-level cluster* and the remaining are *bottom-level clusters*. The *cost of opening* a cluster $\mathcal{K}_i$ is $C_i^o$ and the *cost of closing* it again is $C_i^c$. We define $C_{\mathcal{K}_i} = C_i^o + C_i^c$. An action $\alpha$ belongs to cluster $\mathcal{K}(\alpha)$.

During the course of troubleshooting we gather *evidence* $\varepsilon^i$ meaning that the first $i$ actions failed to solve the problem, and we have by assumption $P(\varepsilon^0) = 1$ because the device is faulty. We also write $\varepsilon^{x:y}$ as shorthand for $\bigcup_{i=x}^{y}\{\alpha_i = no\}$. $\mathcal{FA}(\varepsilon)$ is the set of *free actions* consisting of all actions (excluding those already performed) from open clusters given evidence $\varepsilon$. $\mathcal{CA}(\varepsilon)$ is the set of *confined actions* consisting of all actions from closed clusters. Note that we have $\mathcal{FA}(\varepsilon) \cup \mathcal{CA}(\varepsilon) \subseteq \mathcal{A}$ and $\mathcal{FA}(\varepsilon) \cap \mathcal{CA}(\varepsilon) = \emptyset$ for all evidence $\varepsilon$. By performing an action $\alpha$ from $\mathcal{CA}(\varepsilon)$ we pay the cost $C_{\mathcal{K}(\alpha)}$ because at this point we are certain that we must both open and close the cluster. In that case $\alpha$ is called an *opening action* (for $\mathcal{K}(\alpha)$), and all remaining actions of $\mathcal{K}(\alpha)$ are *released* by removing them from $\mathcal{CA}(\varepsilon)$ and adding them to $\mathcal{FA}(\varepsilon)$. The *conditional cost* $C_\alpha(\varepsilon)$ of an action $\alpha$ given evidence $\varepsilon$ is given by $C_\alpha + C_{\mathcal{K}(\alpha)}$ if $\alpha \in \mathcal{CA}(\varepsilon)$ and by $C_\alpha$ if $\alpha \in \mathcal{FA}(\varepsilon)$.

Throughout this paper we uphold the following simplifying assumptions about the model:

**1** (The single fault assumption). *Initially the problem is known to exist and it is due to the presence of a single fault from $\mathcal{F}$.*

**2** (The idempotent action assumption). *Repeating a failed action will not fix the problem.*

**3** (The carefulness assumption). *By performing an action or testing the system, we never introduce new faults.*

**4** (The independent actions assumption). *Different actions address different faults.*

**5** (The costless system test assumption). *Checking whether the problem still exists after performing an action can be done at a negligible cost.*

**6** (The inside information assumption). *All clusters have inside information.*

Due to the single-fault assumption we may compute the *repair probability* of an action given evidence $\varepsilon$ as $P(\alpha = yes|\varepsilon) = \sum_{f \in \mathcal{F}} P(\alpha = yes|f) \cdot P(f|\varepsilon)$. In a few places we shall abbreviate $P(\alpha = yes|\varepsilon)$ with $P(\alpha|\varepsilon)$. Due to the independent actions assumption $P(\alpha)/P(\beta) = P(\alpha|\varepsilon)/P(\beta|\varepsilon)$ for all evidence $\varepsilon$ not involving $\alpha$ or $\beta$. We shall therefore abbreviate the initial repair probability $P(\alpha = yes)$ as $P_\alpha$.

A *troubleshooting sequence* is a sequence of actions $s = \langle \alpha_1, \ldots, \alpha_n \rangle$ prescribing the process of repeatedly performing the next action until the problem is fixed or the last action has been performed. We shall write $s[k, m]$ for the subsequence $\langle \alpha_k, \ldots, \alpha_m \rangle$ and $s(k, m)$ for the subsequence $\langle \alpha_{k+1}, \ldots, \alpha_{m-1} \rangle$. The index of an opening action in a troubleshooting sequence s is called an *opening index*, and the *set of all opening indices* for s is denoted $\mathcal{Z}$ with $\mathcal{Z} \subseteq \{1, \ldots, n\}, |\mathcal{Z}| = \ell$. To measure the quality of a given sequence we use the following definition.

**Definition 1.** *Let* $s = \langle \alpha_1, \ldots, \alpha_n \rangle$ *be a troubleshooting sequence. Then the* expected cost of repair (ECR) *of* s *is given by*

$$\mathrm{ECR}(s) = \sum_{i=1}^{n} P(\varepsilon^{i-1}) \cdot C_{\alpha_i}(\varepsilon^{i-1}).$$

Formally, our optimization problem is to find a troubleshooting sequence with minimal ECR. Without cost clusters, the problem is easily solved due to the theorem below.

**Theorem 1** (Kadane and Simon (1977)). *Let* s = $\langle \alpha_1, \ldots, \alpha_n \rangle$ *be a troubleshooting sequences in a model without cost clusters. Then* s *is optimal if and only if*

$$\frac{P_{\alpha_i}}{C_{\alpha_i}} \geq \frac{P_{\alpha_i}}{C_{\alpha_i}} \qquad \text{for } i \in \{1, \ldots, n-1\} \ .$$

If costs are not conditional and actions are independent, the lemma below leads directly to the "P-over-C" algorithm.

**Lemma 1** (Jensen et al. (2001)). *Let* s *be a troubleshooting sequence and let* $\alpha_x$ *and* $\alpha_{x+1}$ *be two adjacent actions in* s. *If* s *is optimal then*

$$C_{\alpha_x}(\varepsilon^{x-1}) + \left(1 - P\left(\alpha_x | \varepsilon^{x-1}\right)\right) \cdot C_{\alpha_{x+1}}(\varepsilon^x) \leq$$
$$C_{\alpha_{x+1}}(\varepsilon^{x-1}) +$$
$$\left(1 - P\left(\alpha_{x+1} | \varepsilon^{x-1}\right)\right) \cdot C_{\alpha_x}(\varepsilon^{x-1}, \alpha_{a+1} = no) \ .$$

With assumption 1, 3, 4, and 5 we may simplify computations and notation somewhat because of the following result.

**Proposition 1.** *Let* s = $\langle \alpha_1, \ldots, \alpha_n \rangle$ *be a troubleshooting sequence. Then the ECR of* s *may be computed as*

$$\text{ECR}(s) = \sum_{i=1}^{n} C_{\alpha_i}(\varepsilon^{i-1}) \cdot \left(1 - \sum_{j=1}^{i-1} P_{\alpha_j}\right) \ ,$$

*where* $1 - \sum_{j=1}^{i-1} P_{\alpha_j} = P(\varepsilon^{i-1})$.

This easy computation of the probabilities can be dated back to (Kalagnanam and Henrion, 1990) and (Heckerman et al., 1995).

Thus, due to our assumptions, we may completely ignore $\mathcal{F}$, P(f), and P($\alpha = yes$|f) once the repair probabilities have been computed. Therefore, we mainly use $P_\alpha$ in the rest of this paper.

Using the set of opening indices $\mathcal{Z}$, we can rewrite the definition of ECR of a sequence s to

$$\text{ECR}(s) = \sum_{i=1}^{n} C_{\alpha_i} \cdot \left(1 - \sum_{j=1}^{i-1} P_{\alpha_j}\right)$$
$$+ \sum_{z \in \mathcal{Z}} C_{\mathcal{K}(\alpha_z)} \cdot \left(1 - \sum_{j=1}^{z-1} P_{\alpha_j}\right) \quad (1)$$

where we have decomposed the terms into those that rely on the cost of performing actions and those that rely on the cost of opening and closing a cluster. We define the *efficiency* of an action $\alpha$ given evidence $\varepsilon$ as ef($\alpha | \varepsilon$) = $P_\alpha / C_\alpha(\varepsilon)$, and we write ef($\alpha$) for the unconditional efficiency $P_\alpha / C_\alpha$. Finally, the *cluster efficiency* of an opening action is cef($\alpha$) = $\frac{P_\alpha}{C_\alpha + C_{\mathcal{K}(\alpha)}}$.

**Lemma 2.** *Let* s = $\langle \alpha_1, \ldots, \alpha_n \rangle$ *be an optimal troubleshooting sequence with opening indices* $z_i \in \mathcal{Z}$. *Then the* $\ell + 1$ *subsequences* s$[\alpha_1, \alpha_{z_1})$, s$[\alpha_{z_i}, \alpha_{z_{i+1}}) \ \forall i \in \{1, \ldots, \ell - 1\}$, *and* s$[\alpha_{z_\ell}, \alpha_n]$ *are ordered with respect to descending efficiency.*

*Proof.* Between opening indices the costs are not conditional, and so we must sort by descending ef($\cdot$) to be optimal. □

We have now established that given the opening index for each cluster, it is a simple task of merging ordered sequences to establish an optimal sequence. The difficult part is to determine the opening indices.

## 3 THE EXTENDED P-OVER-C ALGORITHM

The standard "P-over-C" algorithm works by sorting the actions based on descending efficiency. The extended algorithm works in a similar manner, but it also considers the efficiency of a cluster: if a cluster is more efficient than all remaining actions and clusters, we should perform some actions from that cluster first.

**Definition 2.** *The* efficiency *of a cluster* $\mathcal{K}$ *is defined as*

$$\text{ef}(\mathcal{K}) = \max_{\mathcal{M} \subseteq \mathcal{K}} \frac{\sum_{\alpha \in \mathcal{M}} P_\alpha}{C_\mathcal{K} + \sum_{\alpha \in \mathcal{M}} C_\alpha}$$

*and the largest set* $\mathcal{M} \subseteq \mathcal{K}$ *that maximizes the efficiency is called the* maximizing set *of* $\mathcal{K}$. *The sequence of actions found by sorting the actions of the maximizing set by descending efficiency is called the* maximizing sequence *of* $\mathcal{K}$.

It turns out that it is quite easy to calculate the efficiency of a cluster. The following result is a slightly more informative version of the one from (Langseth and Jensen, 2001):

**Lemma 3.** *Let* $\mathcal{K}$ *be a cluster. Then the maximizing set* $\mathcal{M}$ *can be found by including the most efficient actions of* $\mathcal{K}$ *until* ef($\mathcal{K}$) *starts decreasing. Furthermore, all actions* $\alpha$ *in the maximizing set* $\mathcal{M}$ *have* ef($\alpha$) $\geq$ ef($\mathcal{K}$) *and all actions* $\beta \in \mathcal{K} \setminus \mathcal{M}$ *have* ef($\beta$) < ef($\mathcal{K}$).

The algorithm is described in Algorithm 1. If $n$ denotes the total number of actions, we can see that line 2 takes at most $O(n \cdot \lg n)$ time. Once the actions have been sorted, line 3-6 takes at most $O(n)$ time. The loop in line 7-20 can be implemented to run in at most $O(n \cdot \lg(\ell + 1))$ time by using a priority queue for the most efficient element of $\mathcal{A}$ and the most efficient element of each cluster. Thus the algorithm has $O(n \cdot \lg n)$ worst case running time. In the next section we prove that the algorithm returns an optimal sequence.

**Algorithm 1** The extended P-over-C algorithm (Langseth and Jensen, 2001)

---
1: **function** EXTENDEDPOVERC($\mathcal{K}, \mathcal{K}_1, \ldots, \mathcal{K}_\ell$)
2:   Sort actions of $\mathcal{K}$ and all $\mathcal{K}_i$ by descending ef($\cdot$)
3:   Calculate ef($\mathcal{K}_i$) and maximizing sets $\mathcal{M}_i$
4:     for all $i \in \{1, \ldots, \ell\}$
5:   Let $\mathcal{K}_{closed} = \{\mathcal{K}_i \mid i \in \{1, \ldots, \ell\}\}$
6:   Let $\mathcal{A} = \{\alpha \mid \alpha \in \mathcal{K}$ or $\alpha \in \mathcal{K}_i \setminus \mathcal{M}_i$ for some $i\}$
7:   Let s $= \langle \rangle$
8:   **repeat**
9:     Let $\beta$ be the most efficient action in $\mathcal{A}$
10:      or cluster in $\mathcal{K}_{closed}$
11:    **if** $\beta$ is an action **then**
12:      Add action $\beta$ to s
13:      Set $\mathcal{A} = \mathcal{A} \setminus \{\beta\}$
14:    **else**
15:      Add all actions of the maximizing set
16:        of cluster $\beta$ to s in order of
17:        descending efficiency
18:      Set $\mathcal{K}_{closed} = \mathcal{K}_{closed} \setminus \{\beta\}$
19:    **end if**
20:  **until** $\mathcal{K}_{closed} = \emptyset$ and $\mathcal{A} = \emptyset$
21:  Return s
22: **end function**
---

**Example 1.** *We consider a model with three clusters, where $\mathcal{K}_\alpha$ is the root cluster and $\mathcal{K}_\beta$ and $\mathcal{K}_\gamma$ are the bottom-level clusters. We have $C_{\mathcal{K}_\beta} = 2$ and $C_{\mathcal{K}_\gamma} = 1$, and the following model parameters:*

|          | P     | C | ef($\cdot$) | cluster           | ef($\mathcal{K}$) |
|----------|-------|---|-------------|-------------------|-------------------|
| $\alpha_1$ | 0.14  | 1 | 0.14        | $\mathcal{K}_\alpha$ |                   |
| $\alpha_2$ | 0.11  | 1 | 0.11        | $\mathcal{K}_\alpha$ |                   |
| $\beta_1$  | 0.20  | 1 | 0.067       | $\mathcal{K}_\beta$  | 0.075             |
| $\beta_2$  | 0.10  | 1 | 0.033       | $\mathcal{K}_\beta$  |                   |
| $\gamma_1$ | 0.25  | 1 | 0.125       | $\mathcal{K}_\gamma$ | 0.15              |
| $\gamma_2$ | 0.20  | 1 | 0.10        | $\mathcal{K}_\gamma$ |                   |

*The maximizing set for $\mathcal{K}_\beta$ is $\{\beta_1, \beta_2\}$ and for $\mathcal{K}_\gamma$ it is $\{\gamma_1, \gamma_2\}$, and from this the cluster efficiencies have been calculated. Algorithm 1 returns the sequence* s $= \langle \gamma_1, \gamma_2, \alpha_1, \alpha_2,, \beta_1, \beta_2 \rangle$ *which has ECR*

ECR (s) $= 2+0.75+0.55+0.41+0.30 \cdot 3+0.10 = 4.71$ .

*If we followed the simple P-over-C algorithm we would get the sequence* s$^2 = \langle \alpha_1, \gamma_1, \alpha_2, \gamma_2, \beta_1, \beta_2 \rangle$ *with ECR*

ECR $(s^2) = 1+0.86 \cdot 2+0.61+0.50+0.30 \cdot 3+0.10 = 4.83$ .

## 4 CORRECTNESS OF THE ALGORITHM

We start with a proof of Lemma 3:

*Proof.* We shall use the fact that for positive reals we have

$$\frac{a+b}{c+d} \otimes \frac{a}{c} \Leftrightarrow \frac{b}{d} \otimes \frac{a}{c}$$

for any weak order $\otimes$ (e.g. $\geq$ and $\leq$). Let $\mathcal{M}$ consist of actions in $\mathcal{K}$ such that ef($\mathcal{M}$) is maximized. Then ef($\mathcal{M}$) equals

$$\frac{\sum_{\alpha \in \mathcal{M}} P_\alpha}{C_\mathcal{K} + \sum_{\alpha \in \mathcal{M}} C_\alpha} = \frac{\sum_{\alpha \in \mathcal{M} \setminus \{\beta\}} P_\alpha + P_\beta}{C_\mathcal{K} + \sum_{\alpha \in \mathcal{M} \setminus \{\beta\}} C_\alpha + C_\beta}$$
$$= \frac{S_P + P_\beta}{S_C + C_\beta} = \frac{P}{C}$$

where $\beta$ is chosen arbitrarily. Let furthermore $\gamma \in \mathcal{K} \setminus \mathcal{M}$. We shall prove

$$\frac{P_\beta}{C_\beta} \geq \frac{P}{C} > \frac{P_\gamma}{C_\gamma}$$

which implies the theorem. We first prove the leftmost inequality. Because ef($\mathcal{M}$) is maximal we have

$$\frac{S_P + P_\beta}{S_C + C_\beta} \geq \frac{S_P}{S_C} \text{ which is equivalent to } \frac{P_\beta}{C_\beta} \geq \frac{S_P}{S_C}$$

which again is equivalent to

$$\frac{P_\beta}{C_\beta} \geq \frac{S_P + P_\beta}{S_C + C_\beta} \ .$$

The second inequality is proved similarly. □

When we look at opening indices we get the following result.

**Lemma 4.** *Let* s $= \langle \ldots, \alpha_x, \alpha_{x+1}, \ldots \rangle$ *be an optimal troubleshooting sequence, and let $\mathcal{Z}$ be the opening indices of* s. *Then*

$$\begin{array}{ll} \text{cef}(\alpha_x) \geq \text{ef}(\alpha_{x+1}) & \text{if } x \in \mathcal{Z}, \alpha_{x+1} \in \mathcal{FA}(\varepsilon^{x-1}) \\ \text{ef}(\alpha_x) \geq \text{cef}(\alpha_{x+1}) & \text{if } \alpha_x \in \mathcal{FA}(\varepsilon^{x-1}), x+1 \in \mathcal{Z} \\ \text{cef}(\alpha_x) \geq \text{cef}(\alpha_{x+1}) & \text{if } x \in \mathcal{Z}, x+1 \in \mathcal{Z} \end{array}$$

*Proof.* Apply Lemma 1 and do some pencil pushing. For example, case 1: $x \in \mathcal{Z}$ and $\alpha_{x+1} \in \mathcal{FA}(\varepsilon^{x-1})$. In this case we have

$$C_{\alpha_x} + C_{\mathcal{K}(\alpha_x)} + \left(1 - P(\alpha_x | \varepsilon^{x-1})\right) \cdot C_{\alpha_{x+1}} \leq$$
$$C_{\alpha_{x+1}} + \left(1 - P(\alpha_{x+1} | \varepsilon^{x-1})\right) \cdot \left(C_{\alpha_x} + C_{\mathcal{K}(\alpha_x)}\right)$$
$$\Updownarrow$$
$$P(\alpha_{x+1} | \varepsilon^{x-1}) \left[C_{\alpha_x} + C_{\mathcal{K}(\alpha_x)}\right] \leq P(\alpha_x | \varepsilon^{x-1}) C_{\alpha_{x+1}}$$
$$\Updownarrow$$
$$\text{ef}(\alpha_{x+1}) \leq \text{cef}(\alpha_x)$$

because $P(\alpha_x) \geq P(\alpha_{x+1}) \Leftrightarrow P(\alpha_x | \varepsilon) \geq P(\alpha_{x+1} | \varepsilon)$ for independent actions. □

**Definition 3.** *Let* $s[x, y]$ *be a subsequence of a troubleshooting sequence* s. *Then the* efficiency *of* $s[x, y]$ *is given by*

$$\mathrm{ef}(s[x, y]) = \frac{\sum_{i=x}^{y} \mathrm{P}_{\alpha_i}}{\sum_{i=x}^{y} \mathrm{C}_{\alpha_i}(\varepsilon^{i-1})}$$

**Definition 4.** *Let* $s = \langle \ldots, \alpha_x, \ldots, \alpha_y, \ldots \rangle$ *be a troubleshooting sequence. If all actions of the subsequence* $s[x, y]$ *belong to the same cluster, we say that the subsequence is* regular. *If furthermore* $s[x, y]$ *is as long as possible while not breaking regularity, we say that the subsequence is a* maximal regular subsequence.

*Remark.* Any troubleshooting sequence can be partitioned into a sequence of regular subsequences, and if all the subsequences are maximal, this partition is unique.

**Lemma 5.** *Let* s *be an optimal troubleshooting sequence, and let* $s[x, x+k]$ *and* $s[y, y+\ell]$ *(with* $y = x+k+1$*) be two adjacent regular subsequences such that* $\mathcal{K}(\alpha_x) \neq \mathcal{K}(\alpha_y)$ *or such that neither* $x$ *nor* $y$ *is an opening index. Then*

$$\mathrm{ef}(s[x, x+k]) \geq \mathrm{ef}(s[y, y+\ell])$$

*Proof.* We consider the sequence

$$s^2 = \langle \ldots, \alpha_{x-1}, \alpha_y, \ldots, \alpha_{y+\ell}, \alpha_x, \ldots, \alpha_{x+k}, \ldots \rangle$$

which is equal to s except that the two regular sequences have been swapped. Since s is optimal we have $\mathrm{ECR}(s) - \mathrm{ECR}(s^2) \leq 0$. Because the subsequences are regular and belong to different clusters or do not contain opening indices, the costs are the same in the two sequences in both s and $s^2$. Therefore, we get that the terms of $\mathrm{ECR}(s) - \mathrm{ECR}(s^2)$ equal

$$\mathrm{C}_{\alpha_x}(\varepsilon^{x-1}) \cdot \left[ \mathrm{P}(\varepsilon^{x-1}) - \mathrm{P}(\varepsilon^{x-1}, \varepsilon^{y:y+\ell}) \right]$$
$$\vdots$$
$$\mathrm{C}_{\alpha_{x+k}}(\varepsilon^{x+k-1}) \cdot \left[ \mathrm{P}(\varepsilon^{x+k-1}) - \mathrm{P}(\varepsilon^{x+k-1}, \varepsilon^{y:y+\ell}) \right]$$
$$\mathrm{C}_{\alpha_y}(\varepsilon^{y-1}) \cdot \left[ \mathrm{P}(\varepsilon^{y-1}) - \mathrm{P}(\varepsilon^{x-1}) \right]$$
$$\vdots$$
$$\mathrm{C}_{\alpha_{y+\ell}}(\varepsilon^{y+\ell-1}) \cdot \left[ \mathrm{P}(\varepsilon^{y+\ell-1}) - \mathrm{P}(\varepsilon^{x-1}, \varepsilon^{y:y+\ell-1}) \right]$$

since the remaining terms cancel out. Now observe that

$$\mathrm{P}(\varepsilon^{x+i-1}) - \mathrm{P}(\varepsilon^{x+i-1}, \varepsilon^{y:y+\ell}) =$$
$$1 - \sum_{j=1}^{x+i-1} \mathrm{P}_{\alpha_j} - \left[ 1 - \sum_{j=1}^{x+i-1} \mathrm{P}_{\alpha_j} - \sum_{j=y}^{y+\ell} \mathrm{P}_{\alpha_j} \right] = \sum_{j=y}^{y+\ell} \mathrm{P}_{\alpha_j}$$

and, similarly,

$$\mathrm{P}(\varepsilon^{y+i-1}) - \mathrm{P}(\varepsilon^{x-1}, \varepsilon^{y:y+i-1}) =$$
$$1 - \sum_{j=1}^{y+i-1} \mathrm{P}_{\alpha_j} - \left[ 1 - \sum_{j=1}^{x-1} \mathrm{P}_{\alpha_j} - \sum_{j=y}^{y+i-1} \mathrm{P}_{\alpha_j} \right] = -\sum_{j=x}^{x+k} \mathrm{P}_{\alpha_j}$$

So $\mathrm{ECR}(s) - \mathrm{ECR}(s^2) \leq 0$ is equivalent to

$$\left[ \sum_{i=x}^{x+k} \mathrm{C}_{\alpha_i}(\varepsilon^{i-1}) \right] \cdot \sum_{j=y}^{y+\ell} \mathrm{P}_{\alpha_j} \leq \left[ \sum_{i=y}^{y+\ell} \mathrm{C}_{\alpha_i}(\varepsilon^{i-1}) \right] \cdot \sum_{j=x}^{x+k} \mathrm{P}_{\alpha_j}$$

which yields the result. □

**Lemma 6.** *There exists an optimal troubleshooting sequence* s *where for each opening index* $x \in \mathcal{Z}$*, there is a maximal regular subsequence* $s[x, x+j]$ *(*$j \geq 0$*) that contains the maximizing sequence for cluster* $\mathcal{K}(\alpha_x)$.

*Proof.* Let s be an optimal troubleshooting sequence, and let $x$ be an opening index. Let $s[x, x+j]$ be a maximal regular subsequence and assume that it does not contain the maximizing set. Then there exists $\alpha_y \in \mathcal{K}(\alpha_x)$ with $y > x+j+1$ such that

$$\mathrm{ef}(\alpha_y) > \mathrm{ef}(s[x, x+j])$$

Observe that the subsequence $s[x, y-1]$ can be partitioned into $m > 1$, say, maximal regular subsequences $s_1, \ldots, s_m$ with $s_1 = s[x, x+j]$. By Lemma 5 we have

$$\mathrm{ef}(\alpha_y) > \mathrm{ef}(s_1) \geq \mathrm{ef}(s_2) \geq \cdots \geq \mathrm{ef}(s_m) \geq \mathrm{ef}(\alpha_y)$$

where the last inequality follows by the fact that $\alpha_y$ is not an opening action (so we avoid $\geq \mathrm{cef}(\alpha_y)$). This situation is clearly impossible. Therefore $s[x, x+j]$ must contain the maximizing set. By Lemma 2, it must also contain a maximizing sequence. □

*Remark.* In the above proof there is a technicality that we did not consider: there might be equality between the efficiency of an action in the maximizing sequence, the efficiency of the maximizing sequence, and one or more free actions. This problem can always be solved by rearranging the actions, and so for all proofs we shall ignore such details for the sake of clarity.

Finally, we have the following theorem:

**Theorem 2.** *Algorithm 1 returns an optimal troubleshooting sequence.*

*Proof.* By Lemma 5 we know that an optimal sequence can be partitioned into a sequence of maximal regular subsequences which is sorted by descending efficiency. If we consider Lemma 6 too, then we know that we should open the clusters in order of highest efficiency and perform at least all actions in their maximizing sequences as computed by Lemma 3. By Lemma 2 we know that the order of actions in the maximizing sequences is the optimal one. By Lemma 5 we also know that all free actions $\alpha$ with $\mathrm{ef}(\alpha) > \mathrm{ef}(\mathcal{K})$ must be performed before opening the cluster, and all free actions with $\mathrm{ef}(\alpha) < \mathrm{ef}(\mathcal{K})$ must be performed after opening the cluster and performing all the actions in its maximizing sequence. □

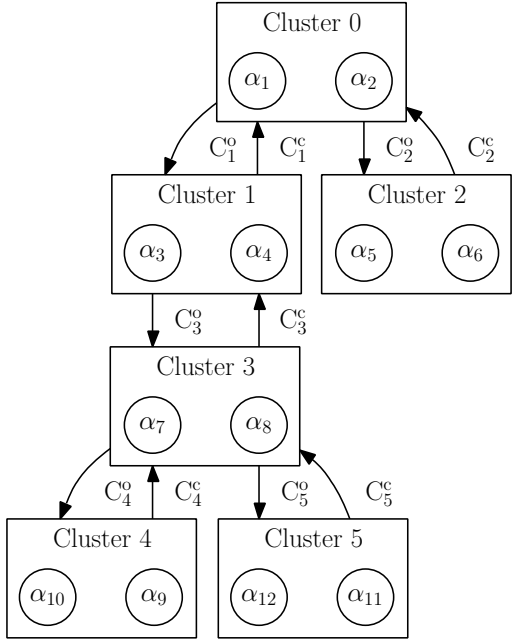

Figure 2: Example of a tree cost cluster model. To open a cluster $\mathcal{K}_i$ when the parent cluster is open we pay the cost $C_i^o$ and to close a cluster given that all children clusters are closed we pay the cost $C_i^c$.

## 5 THE TREE CLUSTER MODEL

In this section we shall investigate an extension of the flat cluster model where the clusters can be arranged as a tree. We call such a model for a *tree cluster model*, and an example is given in Figure 2. In the tree cluster model, the ECR does not admit the simple decomposition of Equation 1. The complication is that several clusters might need to be opened before performing an action in a deeply nested cluster. We therefore call troubleshooting sequences in the tree cluster model for *tree troubleshooting sequences*. Unfortunately, it is easy to construct examples that show that Algorithm 1 will not yield optimal tree troubleshooting sequences. Therefore, we shall present a new algorithm that solves the tree cluster model in $O(n \cdot \lg n)$ time.

First we need some additional definitions. The conditional cost $C_\alpha(\varepsilon)$ of $\alpha \in \mathcal{K}_i$ will now depend on how many clusters that have been opened on the path from the root $\mathcal{K}$ to $\mathcal{K}_i$. We therefore let $\mathcal{AK}(\mathcal{K}_i | \varepsilon)$ denote the *set of ancestor clusters* that needs to be opened on the path from the root $\mathcal{K}$ to $\mathcal{K}_i$ given evidence $\varepsilon$. We then define

$$C_{\mathcal{K}_i}(\varepsilon) = \sum_{\mathcal{K} \in \mathcal{AK}(\mathcal{K}_i | \varepsilon)} C_\mathcal{K}, \quad C_\alpha(\varepsilon) = C_\alpha + C_{\mathcal{K}(\alpha)}(\varepsilon)$$

Given this, Definition 1 is still valid for tree troubleshooting sequences.

A single action is called an *atomic action*. A *compound action* consists of opening a cluster $\mathcal{K}$ and a sequence of actions in which each action may be either atomic or compound. Note that we shall usually not distinguish syntactically between atomic and compound actions. Also note that a compound action corresponds to a subsequence where the first action is an opening action, and the efficiency of a compound action is simply defined as the efficiency of the corresponding subsequence. If $\mathcal{T}$ is a tree cluster model and $\mathcal{K}$ is an arbitrary cluster in $\mathcal{T}$, then the *subtree model induced by $\mathcal{K}$*, denoted $\mathcal{T}_\mathcal{K}$, is a new tree cluster model containing exactly the clusters in the subtree rooted at $\mathcal{K}$, and with $\mathcal{K}$ as the open root cluster. If the induced subtree model is a flat cluster model, we call it a *flat subtree model*.

**Definition 5.** *Let $\mathcal{T}_\mathcal{K} = \{\mathcal{K}, \mathcal{K}_1, \ldots, \mathcal{K}_\ell\}$ be a flat subtree model. Then the* absorbtion *of $\mathcal{K}_1, \ldots, \mathcal{K}_\ell$ into $\mathcal{K}$ is a new cluster $\mathcal{K}^\uparrow$ containing*

1. *for each child cluster $\mathcal{K}_i$, a compound action induced by the maximizing sequence for $\mathcal{K}_i$, and*

2. *all remaining actions from $\mathcal{K}, \mathcal{K}_1, \ldots, \mathcal{K}_\ell$.*

Note that in $\mathcal{K}^\uparrow$ all the actions in a child cluster $\mathcal{K}_i$ that are not contained in the newly generated compound action will have a lower efficiency than the compound action for $\mathcal{K}_i$.

**Definition 6.** *Let $\mathcal{T}$ be a tree cluster model, and let $\mathcal{K}$ be any cluster in $\mathcal{T}$. Then $\mathcal{T}_\mathcal{K}$ may be transformed into a single cluster $\mathcal{K}^\uparrow$ by repeated absorbtion into the root cluster of flat subtree models. The resulting cluster $\mathcal{K}^\uparrow$ is called the* model induced by absorbtion into $\mathcal{K}$.

*Remark.* By construction, the compound actions in a model induced by absorbtion into the root cluster $\mathcal{K}$ will only contain actions from the subtrees rooted at a child of $\mathcal{K}$.

With these definitions we can now present Algorithm 2. The algorithm works in a bottom-up fashion, basically merging leaf clusters into their parents (absorbtion) until the tree is reduced to a single cluster. Then an optimal sequence is constructed by unfolding compound actions when they are most efficient.

The algorithm can be made to run in $O(n \cdot \lg n)$ time by the following argument. Sort the actions of all clusters in the tree $\mathcal{T}$—this takes at most $O(n \cdot \lg n)$ time. During absorbtion, it is important to avoid merging all actions of the child clusters into the parent cluster. Instead, we merge only the compound actions into the parent cluster (takes $O(\ell \cdot \lg n)$ time overall), and create a priority queue holding the most efficient remaining action of each child cluster. When creating a compound action for a parent cluster, we then use actions from the priority queue as needed, and update the priority queue whenever an action is taken out. Therefore, creating all the compound actions can never take more than $O(n \cdot \lg \ell)$ time. As the absorbtion process moves towards the root, we are forced to merge priority queues from different subtrees. A simple induction argument can establish that it takes at most $O(\ell \cdot \lg \ell)$ time to merge all these priority queues.

**Algorithm 2** The bottom-up P-over-C algorithm
```
function BOTTOMUPPOVERC(T)
    Input: a cluster tree T with root K
    Compute the model K↑ induced by absorbtion
        into K (see Definition 6)
    Let s = ⟨⟩
    while K↑ ≠ ∅ do
        Let β be the most efficient action in K↑
        if β is an atomic action then
            Add action β to s
        else
            Add all actions of β to s in the order
            prescribed by β
        end if
        Set K↑ = K↑ \ {β}
    end while
    Return s
end function
```

In the following we shall prove that Algorithm 2 computes an optimal tree troubleshooting sequence. The first two lemmas are minor generalizations of previous lemmas, and the proofs are almost identical.

**Lemma 7.** *Lemma 2 generalizes to tree troubleshooting sequences.*

**Lemma 8.** *Lemma 5 generalizes to subsequences of actions that consists of (i) only free actions, or (ii) actions from the same subtree.*

Next we shall investigate the special properties of the compound actions generated by the absorbtion process.

**Definition 7.** *Let $\mathcal{T}$ be a tree cluster model, and let $\mathcal{K}$ be any non-leaf cluster in $\mathcal{T}$. A* maximizing compound action $\hat{\alpha}$ *for $\mathcal{K}$ in $\mathcal{T}$ is defined as any most efficient compound action in the model induced by absorbtion into $\mathcal{K}$.*

**Lemma 9.** *Let $\mathcal{T}$ be a tree cluster model, and let $\mathcal{K}$ be any non-leaf cluster in $\mathcal{T}$. Let $\mathcal{T}_\mathcal{K}$ be the subtree model induced by $\mathcal{K}$, and let $\hat{\alpha}$ be a maximizing compound action for $\mathcal{K}$ in $\mathcal{T}$. Then*

$$\mathrm{ef}(\hat{\alpha}) \geq \mathrm{ef}(\beta)$$

*where $\beta$ is any possible compound action in $\mathcal{T}_\mathcal{K}$ not including actions from $\mathcal{K}$.*

*Proof.* We proceed by induction. Basis is a flat cluster model $\mathcal{T} = \{\mathcal{K}, \mathcal{K}_1, \ldots, \mathcal{K}_\ell\}$ with compound actions $\hat{\beta}_1, \ldots, \hat{\beta}_\ell$ of $\mathcal{K}$ and $\hat{\alpha} = \max_i \hat{\beta}_i$. Let $\beta$ be any compound action including actions from clusters in $\mathcal{T} \setminus \{\mathcal{K}\}$, and assume that $\mathrm{ef}(\beta) > \mathrm{ef}(\hat{\alpha})$. We shall use the fact

$$\min_i^n \frac{P_i}{C_i} \leq \frac{\sum_i^n P_i}{\sum_i^n C_i} \leq \max_i^n \frac{P_i}{C_i} \quad (2)$$

(which is also known as Cauchy's third inequality). Then by Equation 2, $\beta$ cannot be formed by any combination of the $\hat{\beta}_i$'s as this would not increase the efficiency. Therefore $\beta$ must be formed by either a strict subset or a strict superset of one of the $\hat{\beta}_i$'s. If $\beta$ is a subset of any $\hat{\beta}_i$, then the maximality of $\hat{\beta}_i$ leads to a contradiction. If $\beta$ is a superset of any $\hat{\beta}_i$, then it will include subsets of actions from a set of clusters with subscripts $\mathcal{I} \subseteq \{1, \ldots, \ell\}$. Let us denote the subsets from each $\mathcal{K}_i$ as $\beta_i$. We then have

$$\mathrm{ef}(\beta) = \frac{\sum_{i \in \mathcal{I}} P_{\beta_i}}{\sum_{i \in \mathcal{I}} C_{\beta_i}} \leq \max_{i \in \mathcal{I}} \frac{P_{\beta_i}}{C_{\beta_i}} \leq \max_{i \in \{1, \ldots, \ell\}} \frac{P_{\hat{\beta}_i}}{C_{\hat{\beta}_i}} = \mathrm{ef}(\hat{\alpha})$$

where the first inequality follows by Equation 2, the second follows by the definition of compound actions formed during absorbtion, and the last equality is by definition of a maximizing compound action. Since the sets $\beta_i$ were chosen arbitrarily, we get a contradiction. Hence in all cases $\mathrm{ef}(\hat{\alpha}) \geq \mathrm{ef}(\beta)$.

Induction step: we assume the Lemma is true for all children $\mathcal{K}_i, \ldots, \mathcal{K}_\ell$ of an arbitrary cluster $\mathcal{K}$ where the children have maximizing compound actions $\hat{\beta}_1, \ldots, \hat{\beta}_\ell$. A similar argument as above then shows that the lemma is true for $\mathcal{K}$ as well. □

**Lemma 10.** *Let $\mathcal{T}$ be a tree cluster model with root cluster $\mathcal{K}$. Then there exists an optimal tree troubleshooting sequence s that contains (as subsequences) all the compound actions of the model induced by absorbtion into $\mathcal{K}$. Furthermore, the compound actions in s are ordered by descending efficiency.*

*Proof.* Let $s = \langle \alpha_1, \ldots, \alpha_x, \ldots, \alpha_{x+k}, \ldots \rangle$ be an optimal tree troubleshooting sequence and let $\alpha_x$ be an opening action, and let $s[x, x+k], k \geq x$ be the sequence of maximal length of actions from the same subtree. Let furthermore $s[x, x+k]$ be the first subsequence that contradicts the lemma, that is, $s[x, x+k]$ does not contain the compound action $\hat{\alpha}$ for the cluster $\mathcal{K}(\alpha_x)$. Then there exists an atomic action $\alpha_y \in \hat{\alpha}$ (with $y > x + k + 1$) such that $\alpha_y \notin s[x, x+k]$. We then have

$$\mathrm{ef}(\alpha_y) > \mathrm{ef}(\hat{\alpha}) > \mathrm{ef}(s[x, x+k])$$

because all atomic actions in a compound action are more efficient than the compound action itself, and because $\hat{\alpha}$ is the most efficient compound action in the subtree rooted at $\mathcal{K}(\alpha_\text{x})$ (Lemma 9). We can then partition the actions between $\alpha_\text{x+k}$ and $\alpha_\text{y}$ into $m > 1$, say, subsequences (of maximal length) $s_1, \ldots, s_m$. If one (or more) of these subsequence is more efficient than $\alpha_\text{y}$, we immediately get a contradiction to optimality of s because such a subsequence can be moved before s$[x, x+k]$ (Lemma 8). So we can assume that all the $m$ subsequences are less efficient than $\alpha_\text{y}$. Then by successive application of Lemma 8 we can decrease the ECR by moving $\alpha_\text{y}$ to position $x+k+1$. However, this again contradicts that s was optimal. Hence s$[x, x+k]$ must contain $\hat{\alpha}$.

By Lemma 8 it follows that the order of the compound actions must be by descending efficiency. □

**Theorem 3.** *Algorithm 5 returns an optimal troubleshooting sequence.*

*Proof.* By Lemma 10 we only need to establish the order of the free actions between compound actions. By Lemma 8 it follows that any compound action is preceeded by more efficient free actions and followed by less efficient free actions. □

## 6 CONCLUSION

We have presented an algorithm, which in $O(n \cdot \lg n)$ time ($n$ being the number of actions) provides an optimal troubleshooting sequence for scenarios where the cost clusters form a tree and have inside information. This is a useful result on its own, but there is more to it.

When evaluating algorithms for troubleshooting, you must distinguish between off-line and on-line activity. If your task is off-line, the time complexity of your algorithm may not be particularly important as long as the result can be stored easily (like for example an optimal action sequence). However, if the decision support system is flexible, it must allow the user to interact with the recommendations and have the system calculate an optimal next action based on alternative information.

Furthermore, for many scenarios you will request on-line calculation of an optimal sequence; for example when the model includes questions and tests. For this kind of scenario, a direct representation of an optimal strategy may require too much space. Therefore, a myopic question heuristic usually relies on optimal sequences of actions calculated on-line.

Finally, our results imply a major improvement for off-line methods like AO* because the search tree can now be extensively pruned. This is because all subtrees that consist entirely of actions can be replaced with a single sequence of actions.

## 7 ACKNOWLEDGEMENTS

We would like to thank the three anonymous reviewers for their excellent feedback. Thanks also go to Sven Skyum for help with Lemma 3.